\PassOptionsToPackage{numbers}{natbib}
\documentclass{article}

\usepackage[preprint]{neurips_2024}

\usepackage[utf8]{inputenc} 
\usepackage[T1]{fontenc}    
\usepackage{hyperref}       
\usepackage{url}            
\usepackage{booktabs}       
\usepackage{amsfonts}       
\usepackage{nicefrac}       
\usepackage{microtype}      
\usepackage{xcolor}         

\usepackage{natbib}

\usepackage{amsmath}
\usepackage{graphicx}
\usepackage{authblk}

\title{How Difficulty-Aware Staged Reinforcement Learning Enhances LLMs' Reasoning Capabilities: A Preliminary Experimental Study}

\begin{document}
\author{
  Yunjie Ji,\quad Sitong Zhao,\quad Xiaoyu Tian,\quad Haotian Wang,\\[0.3em]
  Shuaiting Chen,\quad Yiping Peng,\quad Han Zhao,\quad Xiangang Li
}


\affil{
    \raisebox{-0.4em}{\includegraphics[height=1.5em]{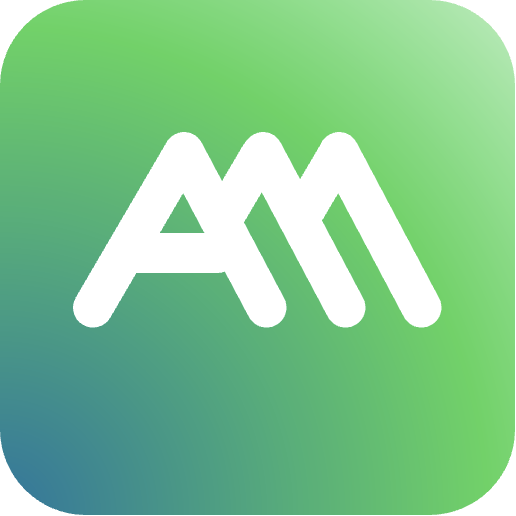}}
    \hspace{0.2em}a-m-team
}
\date{}

\maketitle

\begin{abstract}
Enhancing the reasoning capabilities of Large Language Models (LLMs) with efficiency and scalability remains a fundamental challenge in artificial intelligence research. This paper presents a rigorous experimental investigation into how difficulty-aware staged reinforcement learning (RL) strategies can substantially improve LLM reasoning performance. Through systematic analysis, we demonstrate that strategically selecting training data according to well-defined difficulty levels markedly enhances RL optimization. Moreover, we introduce a staged training methodology, progressively exposing models to increasingly challenging tasks, further amplifying reasoning capabilities. Our findings reveal significant cross-domain benefits when simultaneously training models on mathematical reasoning and code generation tasks. Notably, our proposed approach enables a 1.5B parameter model to achieve an accuracy of 42.3\% on the AIME-2024 benchmark, 89.5\% on the MATH-500 benchmark. These results underscore the efficacy of our method in advancing the reasoning proficiency of LLMs.
We will open-source our datasets on  \href{https://github.com/a-m-team/a-m-models}{GitHub}\footnote{\url{https://github.com/a-m-team/a-m-models}} and  \href{https://huggingface.co/a-m-team}{Hugging Face}\footnote{\url{https://huggingface.co/a-m-team}}.

\end{abstract}

\section{Introduction}

Enhancing the reasoning capabilities of Large Language Models (LLMs) while maintaining scalability and efficiency remains a key challenge in modern AI research. While contemporary LLMs, such as GPT-4 \cite{openai2023gpt4}, Claude 3 \cite{anthropic2024claude3}, and DeepSeek-R1 \cite{deepseek2024r1}, have demonstrated impressive performance across various tasks, further enhancing their reasoning abilities in specialized domains like mathematical reasoning and code generation is still an ongoing challenge \cite{yao2023treeofthought, shinn2023reflexion}. Recently, several promising strategies have emerged, including sophisticated prompting techniques like Chain-of-Thought (CoT) \cite{wei2022chain} and Test-Time Scaling methods \cite{wang2023selfconsistency}, alongside reinforcement learning (RL)-based approaches designed to optimize reasoning capabilities directly \cite{ouyang2022rlhf, silver2023grpo}.

Test-time scaling strategies have shown notable success in improving reasoning during inference by using extended prompts or iterative processes, without retraining the model parameters. Techniques like self-consistency voting, extended chain-of-thought prompts, and Tree-of-Thought search \cite{yao2023treeofthought, zhou2023thinktwice, tian2025thinktwiceenhancingllm} help models explore multiple potential solutions and reflect upon reasoning paths, thereby reducing logical errors and enhancing overall accuracy. However, these methods primarily build upon the model's existing capabilities, and do not address the underlying optimization of reasoning strategies.

Reinforcement learning, on the other hand, offers a more direct approach to improving reasoning performance by explicitly rewarding desired reasoning outcomes. Models like DeepSeek-R1 have demonstrated the potential of RL to significantly improve performance on tasks such as mathematical problem-solving and code generation \cite{deepseek2024r1, kimik2024k1.5}. However, applying RL to optimize reasoning performance comes with its own set of challenges, including the selection of suitable training data, managing the length of reasoning chains, and the quality of the initial baseline model.

In this paper, we focus on addressing these challenges through a systematic investigation of the factors influencing reinforcement learning-based reasoning optimization. Our experiments reveal that the careful selection of appropriately difficult training data, combined with a staged RL training approach, can significantly improve model performance. Specifically, we find that carefully selecting training data with appropriate difficulty levels, combined with a staged training approach, significantly improves model performance. This strategy leads to a performance improvement of 5.6\% on the MATH-500 benchmark and 13.4\% on the AIME-2024 benchmark compared to the base model.

Additionally, we demonstrate the benefits of mixing mathematical reasoning and code generation tasks within the same training process. By training the model on both types of tasks simultaneously, we observe improvements across both domains. Specifically, the performance on the AIME-2024 benchmark improved by 7.8\%, the MATH-500 benchmark improved by 5.2\%, and LiveCodeBench saw a 3.5\% increase, compared to the base model.

Our contributions in this paper are as follows:

\begin{itemize} 
\item We provide a detailed experimental analysis of how training data difficulty and staged training impact reinforcement learning optimization for reasoning tasks. 
\item We demonstrate the effectiveness of mixed-domain training, specifically combining mathematical reasoning and code generation, leading to significant performance improvements across both domains. 
\item We empirically validate our findings by showing a 13.4\% improvement on AIME-2024 and a 5.5\% improvement on MATH-500. \end{itemize}

Our findings offer practical guidance for applying reinforcement learning to enhance reasoning capabilities in LLMs and contribute to the ongoing research on scaling models for complex, multi-domain reasoning tasks.

\section{Related work}
Since the release of DeepSeek R1\cite{deepseek2024r1}, a growing number of research groups and open-source communities have attempted to replicate its remarkable reasoning capabilities. These replication efforts vary in scale and methodology, but collectively provide a valuable understanding of R1's training pipeline and reasoning behavior. In this section, we review key replication efforts and related studies, focusing on whether they provide publicly accessible reports or technical documentation, as well as how their outcomes compare with the original R1 model.

Open-R1\cite{openr1} by Hugging Face is one of the most comprehensive and transparent replication initiatives. The project openly shares its implementation and datasets through GitHub and aims to fully reproduce the three-stage training pipeline of DeepSeek-R1: (1) R1-Distill for supervised fine-tuning via distilled reasoning data, (2) R1-Zero for zero-SFT reinforcement learning using self-generated rewards, and (3) the full multi-stage training involving SFT and RL. Open-R1 maintains detailed project documentation, datasets such as OpenR1-Math-220k, and benchmark results that show competitive performance compared to R1.

OpenThinker-32B\cite{openthinker}, developed by researchers from Stanford and UC Berkeley, demonstrates that with only 114k carefully filtered reasoning samples, it is possible to fine-tune Qwen2.5-32B to match or even surpass the performance of DeepSeek-R1's 32B distilled model. Notably, OpenThinker-32B outperforms R1 on benchmarks such as MATH500 while using just 14\% of the training data. 

The SimpleRL-Reason\cite{simplerlreason} project from HKUST investigates the capability of small models trained with reinforcement learning to replicate R1-Zero-style behavior. Without any supervised pretraining, the authors apply PPO to fine-tune Qwen2.5-Math-7B using only 8K math problems. The resulting model shows strong performance across various math reasoning tasks, outperforming baseline supervised fine-tuned models.

TinyZero\cite{tinyzero}, from UC Berkeley, focuses on low-cost replication of R1-Zero-like abilities using a lightweight reinforcement learning setup. The authors train a 3B model with less than \$30 in compute budget, and demonstrate the emergence of self-corrective reasoning behavior on arithmetic and logic puzzles. 

The Simple-GRPO\cite{simplegrpo} project from Fudan University proposes an extremely lightweight implementation of DeepSeek's Group Relative Policy Optimization (GRPO) algorithm. With only ~300 lines of core code and training conducted on minimal hardware, the project demonstrates that "Aha moments" in R1-Zero-style training can be replicated at extremely low cost.

Datawhale-R1\cite{datawhaler1}, developed by the Datawhale community, offers a Chinese-language tutorial and implementation for replicating R1-Zero behavior. While primarily intended for educational purposes, the project documents the full training process on a local Qwen-7B model, and reports trends consistent with those observed in larger replication efforts.

LIMO (Less Is More for Reasoning)\cite{limo} proposes an alternative to large-scale replication by demonstrating that only 817 high-quality samples are sufficient to fine-tune a Qwen-7B model to state-of-the-art performance across many math reasoning benchmarks. Their paper reports results such as 94.8\% accuracy on MATH500 and 57.1\% on AIME24, significantly outperforming previous models.

Dr. GRPO \cite{drgrpo} analyzes the design and biases of the R1-Zero training framework. The authors identify issues such as length bias in GRPO optimization and propose Dr. GRPO, a debiased version of the algorithm. Their 7B-scale model achieves 43.3\% on AIME24, setting a new SOTA among open-source R1-Zero-style replications.

DeepScaleR\cite{deepscaler2025} introduces a novel approach to scaling reinforcement learning by fine-tuning a 1.5B model with 40,000 math problems, achieving state-of-the-art performance. Their model surpasses OpenAI's O1-Preview with a 43.1\% Pass@1 accuracy on AIME2024, demonstrating significant improvements over previous models. The authors show that efficient scaling can be achieved by reducing computational costs while maintaining high performance, thanks to a distilled model and iterative context lengthening

In summary, these replication projects have collectively shed light on the reproducibility, efficiency, and limitations of R1-style training. By providing open-source code, high-quality datasets, and technical documentation, they offer valuable resources for future work on reasoning-capable language models.

\section{Data Selection}

\subsection{Data Sources}
The training data used in our reinforcement learning  is primarily drawn from high-quality, math- and code-centric datasets such as \texttt{numina-math}\cite{numina_math_datasets} and \texttt{PRIME}\cite{cui2025process}. We follow a meticulous data curation process, inspired by methodologies described in prior work, please refer to \cite{zhao202514millionopensourcedistilled} for more details. 

\subsection{Difficulty Scoring}
To assess the intrinsic difficulty of each data sample, we evaluate the performance of multiple reasoning models on the math and code problems. Specifically, we run \texttt{DeepSeek-R1-Distill-Qwen-1.5B}, \texttt{DeepSeek-R1-Distill-Qwen-7B}, \texttt{DeepSeek-R1-Distill-Qwen-32B} and \texttt{DeepSeek-R1} across the datasets multiple times. Each model’s \textit{pass rate} on individual problems is recorded and used as an approximate measure of difficulty.

To enhance reliability, we use a verifier to check each answer. For math problems, we first extract the answer part from the model's response and use rules to parse the content within the \verb|\boxed{}| expressions. We then compare this extracted content with the ground truth answer and use the Hugging Face Math Verify tool\cite{huggingface2025mathverify} to check whether the two match. Any format inconsistencies or incorrect answers are assigned a score of zero.

\[
\text{score\_math} =
\begin{cases}
0, & \text{if model's response format is incorrect}, \\
0, & \text{if model's answer is incorrect}, \\
1, & \text{if model's answer is correct}.
\end{cases}
\]

For programming problems, we execute the model-generated code within a secure sandboxed environment. The outcome is quantified using the ratio of test cases passed to the total number of test cases: 

\[
\text{score\_code} =
\frac{\texttt{\#\\passed\_test\_case}}{\texttt{\#\\test\_cases}}
\]

Importantly, we discard any samples where the \texttt{DeepSeek-R1} model achieves a pass rate of zero, as such instances are either too difficult or unverifiable due to limitations of the verifier.

\subsubsection{Data Filtering Strategy}
\label{data_difficulty_level_definition}
Our data selection process is based on three distinct difficulty levels, which are designed to conduct comprehensive experiments.

\paragraph{Difficulty Level 1:} 
This subset includes data samples where \texttt{DeepSeek-R1-Distill-Qwen-1.5B} achieves a pass rate strictly between 0 and 1 (i.e., pass rate $\in$ (0, 1)). These samples are considered moderately challenging, as the model is partially successful, making them valuable for training on problems that are neither trivial nor unsolvable.

\paragraph{Difficulty Level 2:} 
This level includes three type of samples:
\begin{itemize}
    \item Samples where \texttt{DeepSeek-R1-Distill-Qwen-1.5B} fails completely (pass rate = 0), but \texttt{DeepSeek-R1-Distill-Qwen-7B} succeeds fully (pass rate = 1). This highlights the advantages of scaling up model size.
    \item Samples where \texttt{DeepSeek-R1-Distill-Qwen-1.5B} fails (pass rate = 0), but \texttt{DeepSeek-R1-Distill-Qwen-7B} partially succeeds (pass rate $\in$ (0, 1)). These samples emphasize the robustness of larger models when smaller models fail.
    \item Samples where both \texttt{DeepSeek-R1-Distill-Qwen-1.5B} and \texttt{DeepSeek-R1-Distill-Qwen-7B} achieve partial success, useful for analyzing common challenges and exploring potential improvements.
\end{itemize}

\paragraph{Difficulty Level 3:} 
This subset consists of difficult tasks where the model is less likely to succeed. These samples are chosen based on the performance of the \texttt{DeepSeek-R1-Distill-Qwen-32B} model:
\begin{itemize}
    \item We retain all samples that the \texttt{32B} model fails to solve, representing the hardest tasks in our dataset.
    \item For the math subset, we retain 50\% of the samples successfully solved by the \texttt{32B} model. For code tasks, we retain only 10\% of such samples. This selective retention ensures that only the most challenging or borderline cases are kept for training.
\end{itemize}

This tiered approach ensures that the resulting dataset includes a balanced distribution of tasks across different difficulty levels, which facilitates subsequent experiments, such as evaluating model performance under varying training data difficulties and conducting staged experiments that gradually increase the difficulty of the training data.

\section{Experiments}
\subsection{Experimental Setup}
\label{exp_settings}

In our experiments, we utilized the GRPO ( Group Relative Policy Optimization) \cite{shao2024deepseekmath} algorithm. The learning rate was set to \(1 \times 10^{-6}\). For each prompt, 16 samples were generated, with the maximum sequence length set to 16k tokens. The training batch size was configured to 128, and on-policy model parameter updates were performed throughout the training process. To control the balance between the policy's exploration and exploitation, we set the KL divergence coefficient to 0.001. We also incorporated an entropy loss term, with the entropy coefficient set to 0.001 to encourage a sufficiently high level of exploration during training. The initial policy model used in our experiments was \texttt{DeepSeek-R1-Distill-Qwen-1.5B}.
For evaluation, we set the generation sequence length limit to 32k tokens, with the temperature set to 0.6 and top\_p set to 0.95. We computed the Pass@1 accuracy as the evaluation metric. For AIME-2024 and MATH-500, we ran the evaluation 16 times, while for LiveCodeBench, we ran it 4 times.

\subsection{Impact of Data Difficulty on Model Performance}
\label{impact_of_data_difficulty}

In reinforcement learning (RL), an essential question is what type of data can facilitate better policy learning. To address this, we selected data based on difficulty level as a key criterion. The difficulty of the data was determined by the pass rate of different models, which served as a metric to categorize the data into three difficulty levels, as described in \ref{data_difficulty_level_definition}.

The initial rewards on the training set for the models were as follows: 0.58, 0.50, and 0.17. These results align with the difficulty trends that were set when constructing the three datasets. The performance of the model on datasets of varying difficulty levels is summarized below.

\begin{figure}[h!]
    \centering
    \includegraphics[width=1.0\textwidth]{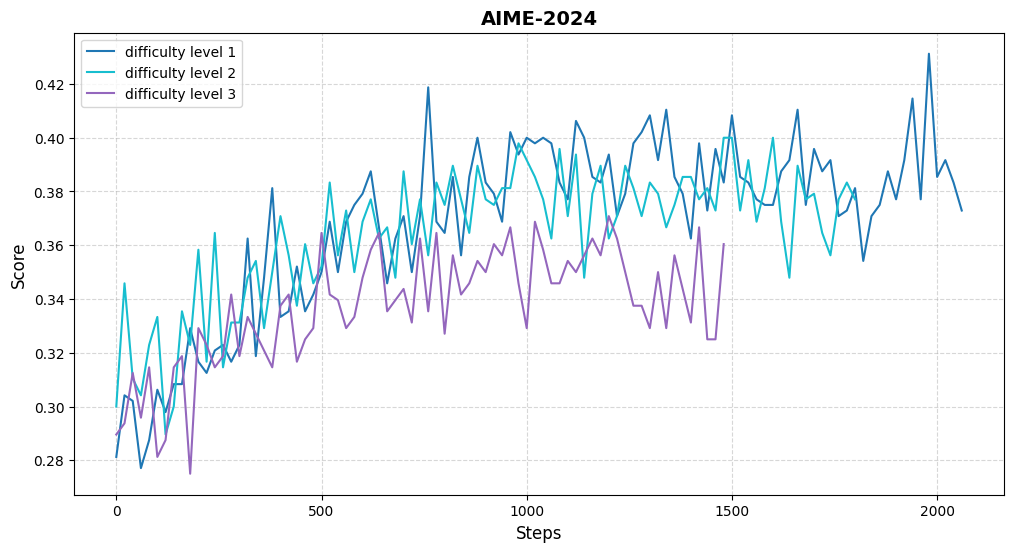}
    \caption{Performance of the model on the AIME-2024 benchmark during training with three different difficulty levels.}
    \label{fig:data-difficulty}
\end{figure}

Figure \ref{fig:data-difficulty} illustrates the performance trajectories of the model across these three difficulty levels. 
The results highlight the importance of selecting appropriately challenging RL training data. Based on our experiments, we found that data with an initial policy pass rate within the range (0, 1)—corresponding to difficulty level 1—yielded the best training results. This dataset offers a moderate challenge for the initial policy. This balance enables the model to generate both correct and incorrect responses during multiple rollouts for the same prompt, thus enhancing the effectiveness of RL training. In contrast, difficulty level 3, which represents highly challenging data, results in some instances where the model continuously fails to generate correct answers, which in turn reduces the model's convergence efficiency and overall performance.

In summary, carefully selecting RL training data based on appropriate difficulty metrics is critical. A moderate difficulty level enhances learning efficiency, balancing the need for adequate challenge against the risk of overwhelming the learning process with overly difficult scenarios.

\begin{figure}[h!]
    \centering
    \includegraphics[width=1.0\textwidth]{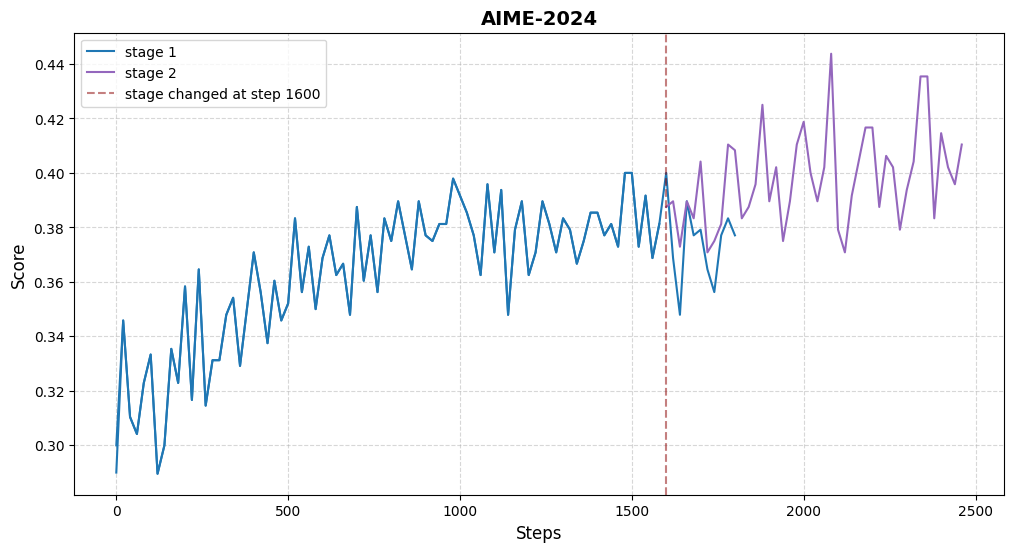}
    \caption{Performance of the model during staged RL training. The plot shows the model's score over time, with stage 1 (blue) and stage 2 (orange). The vertical dashed line indicates the point (step 1600) where the model transitioned from stage 1 to stage 2, resulting in performance improvements.}
    \label{fig:staged_training}
\end{figure}

\subsection{Staged RL Training}
\label{staged_rl_training}
There are several reasons why we may wish to adopt a staged approach in reinforcement learning (RL) training, such as reducing training costs or further enhancing the model’s performance. After conducting a series of experiments, we implemented the following staged RL training strategy. 

\begin{figure}[h!]
    \centering
    \includegraphics[width=1.0\textwidth]{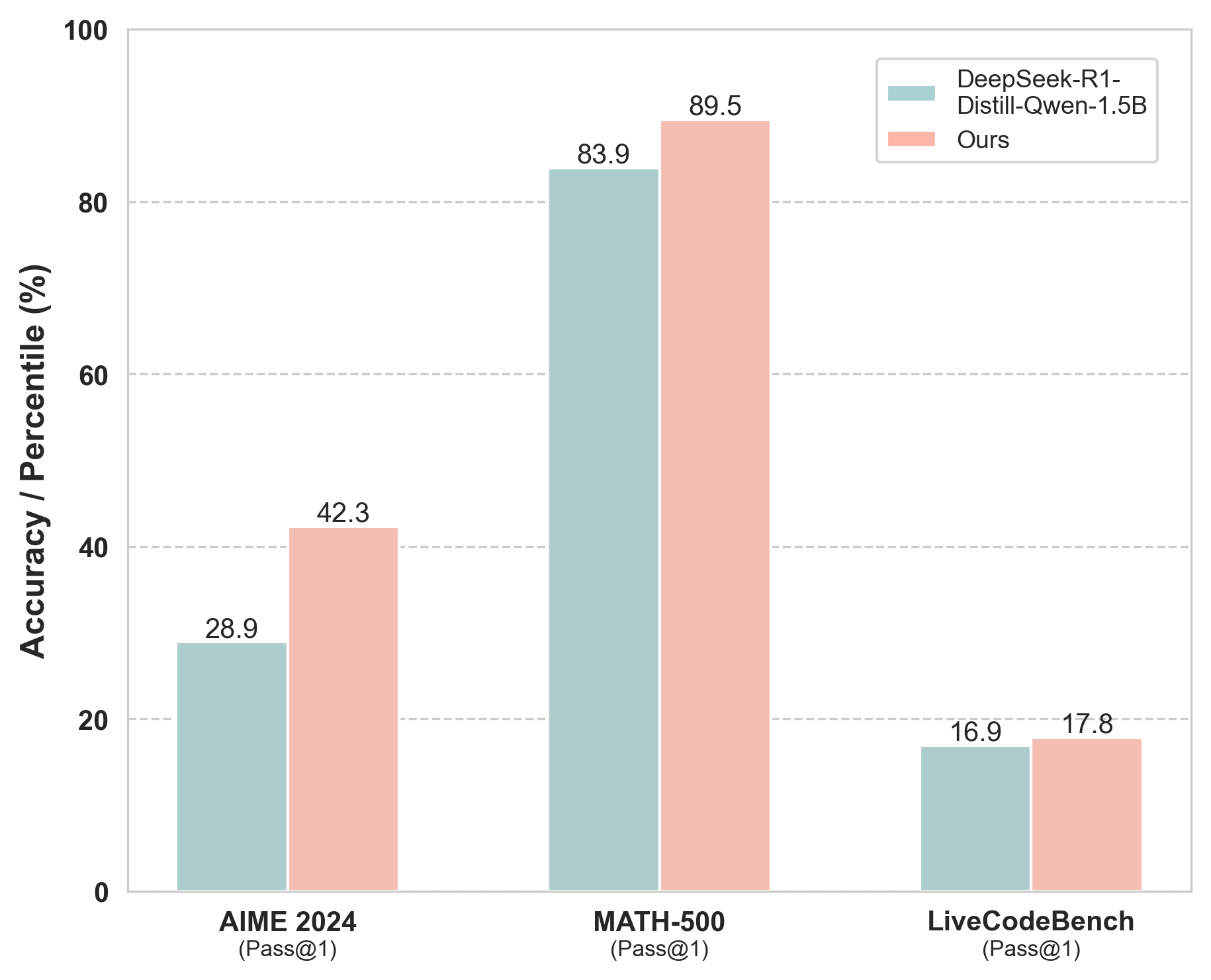}
    \caption{Our model, trained using a two-stage reinforcement learning (RL) approach, shows significant performance improvements on two mathematics-related benchmarks, AIME-2024 and MATH-500. However, due to the absence of code-related training data, its performance on LiveCodeBench is essentially the same as that of the base model.}
    \label{fig:staged2}
\end{figure}

In the first stage, we trained the model using data from difficulty level 2, following the training configurations outlined in \ref{exp_settings}. Once the model’s performance plateaued, we transitioned to the second stage of training. During this phase, we chose to use difficulty level 3 data. Additionally, we adjusted the maximum sequence length for the rollouts to 24k. 

Increasing the maximum sequence length led to a rapid increase in the length of the responses generated by the policy. However, a portion of these responses were truncated due to the length limitation. Providing negative feedback for these truncated responses is not ideal, as it could result in unfairly penalizing the model. Therefore, we opted not to compute the loss for these truncated samples. Furthermore, we decided to exclude the entropy loss from this phase of training.

When comparing this staged approach to the continuous training in the first stage, we observed that switching to the second stage resulted in sustained performance improvements, demonstrating the effectiveness of this strategy in further enhancing model capabilities.

\subsection{Simultaneous Training on Mathematics and Code}


To improve the model's capabilities in various domains, such as mathematics, programming, and general problem-solving, it is essential to include diverse types of prompts in the training dataset, along with corresponding reward rules or reward models. The distribution and difficulty of these different data types significantly impact the model’s performance. 

In our preliminary experiments, we explored training the model on both mathematical and programming tasks simultaneously. Our findings suggest that simply mixing these two types of training data can lead to improvements in both mathematical and programming proficiency. Specifically, we selected difficulty level 1 data for mathematics and difficulty level 3 data for coding, with a ratio of approximately 2.1:1. The mixing strategy employed involved incorporating both mathematical and coding prompts within the same batch during training. 

\begin{figure}[h!]
    \centering
    \includegraphics[width=1.0\textwidth]{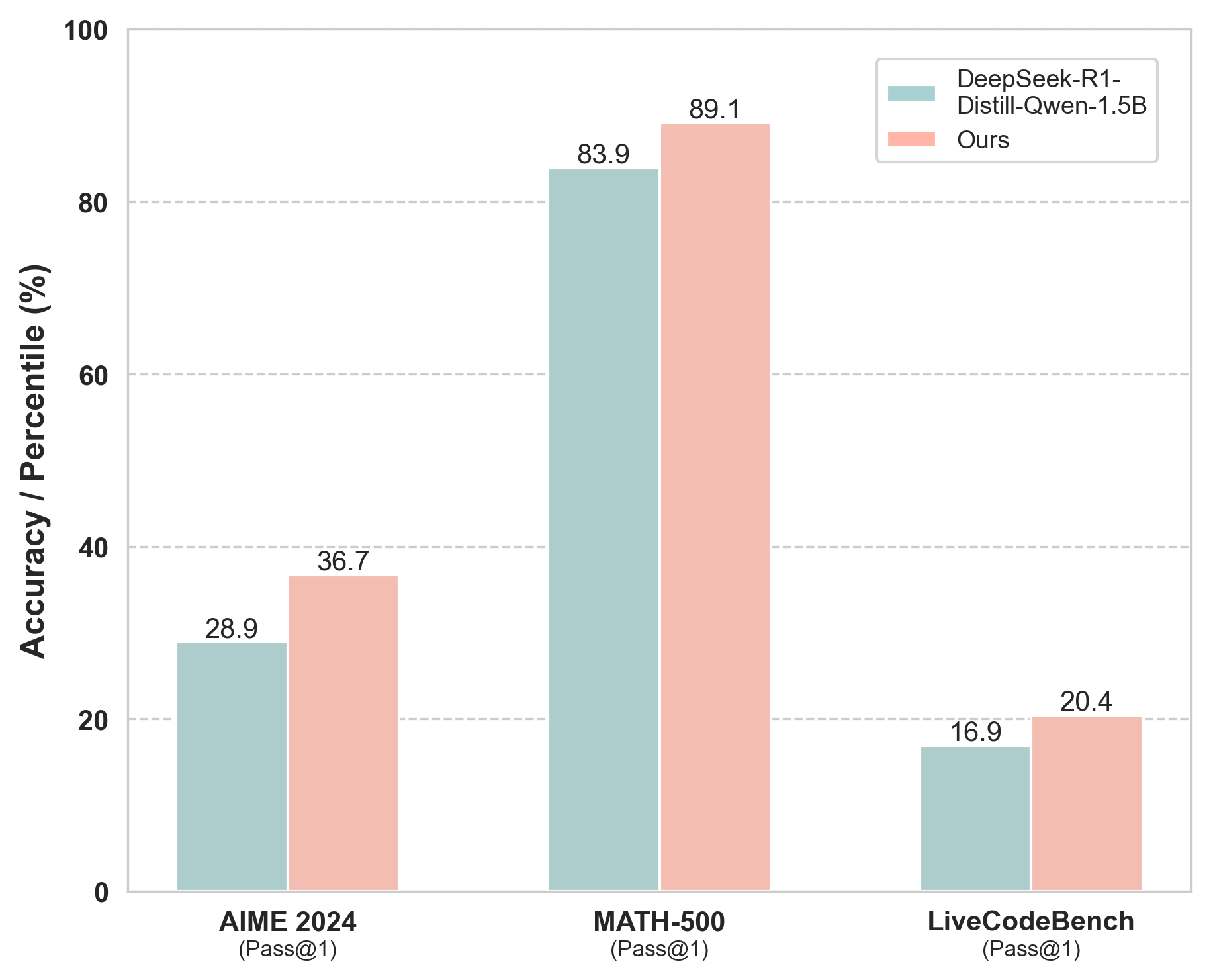}
    \caption{Performance comparison of models across different benchmarks: AIME-2024, MATH-500, and LiveCodeBench. Our model is simultaneously trained on math and code data.}
    \label{fig:math_code}
\end{figure}

Figure \ref{fig:math_code} presents the experimental results, demonstrating the effectiveness of this approach. Based on the conclusions drawn from \ref{impact_of_data_difficulty}, we hypothesize that incorporating difficulty level 1 coding data into the training mix would yield even more favorable outcomes, as it strikes a balance between challenge and learning efficiency.

\section{Conclusion and Limitations}

In this paper, we have explored the potential of using reinforcement learning (RL) to enhance the reasoning capabilities of Large Language Models (LLMs), specifically focusing on the careful selection of training data and staged training strategies. Our experiments demonstrated that by selecting appropriately challenging data and incorporating staged training, we can significantly improve the performance of LLMs on reasoning tasks. Notably, the performance gains were substantial on benchmarks such as MATH-500 and AIME-2024, with improvements of 5.6\% and 13.4\%, respectively. Furthermore, we showed that mixing mathematical reasoning and code generation tasks during training results in cross-domain improvements, providing strong evidence for the benefits of multi-domain training.

These findings provide practical guidance for future work on applying reinforcement learning to optimize reasoning in LLMs and highlight promising avenues for scaling models to handle more complex, multi-domain tasks.

However, despite the promising results, there are some limitations to our approach. Firstly, the selection of appropriate difficulty levels for training data is inherently dependent on the performance of the initial models, which is costly. Further research is needed to refine these difficulty scoring methods and adapt them to a broader range of tasks. Secondly, while the staged RL training approach showed improvements, it is important to note that the performance gains were not uniform across all domains.  While our staged approach reduced training costs, it is still computationally intensive, and further optimization could help reduce the resource requirements.

Lastly, the reliance on the quality of data and the underlying models poses another challenge. Ensuring the quality of training data, especially when dealing with complex reasoning tasks, remains a critical factor. Future work could focus on automating the data curation process and improving the robustness of the RL training pipeline to better handle noisy or ambiguous data.


\bibliographystyle{plainnat}
\bibliography{reference}

\appendix




\end{document}